\documentclass[11pt]{article}
\usepackage{amsmath,amssymb,amsthm,hyperref,multirow}
\usepackage{enumitem}

\usepackage{textcomp}

\usepackage{booktabs}

\usepackage{color}
\usepackage{tikz}

\newsavebox{\measurebox}
\usepackage[numbers,sort&compress]{natbib}
\usepackage{authblk}
\usepackage{algorithm,algorithmic}

\usepackage{dsfont, physics,stmaryrd,tabu}
\usepackage{enumitem}
\usepackage{mathtools}

\theoremstyle{plain}
\newtheorem{theorem}{Theorem}[section]
\newtheorem*{theorem*}{Theorem}
\newtheorem{lemma}[theorem]{Lemma}

\theoremstyle{definition}

\theoremstyle{remark}

\usepackage{authblk}

\numberwithin{equation}{section}
\numberwithin{algorithm}{section}
\numberwithin{figure}{section}
\numberwithin{table}{section}
\numberwithin{theorem}{section}

\newcommand{\R}{\mathbb{R}}
\newcommand{\E}{\mathbb{E}}
\newcommand{\Expect}[1]{\mathbb{E}\left[ #1 \right]}
\usepackage{array}
\newcolumntype{P}[1]{>{\centering\arraybackslash}p{#1}}

\usepackage{pgfplots}
\usepgfplotslibrary{fillbetween}
\usepackage{pgfplots}
\usepackage{filecontents}

\definecolor{warm1}{rgb}{1.0, 0.75, 0.0} 
\definecolor{warm2}{rgb}{1.0, 0.49, 0.0} 
\definecolor{warm3}{rgb}{1.0, 0.01, 0.24} 

\definecolor{cold1}{rgb}{0.4, 0.6, 0.8} 
\definecolor{cold2}{rgb}{0.0, 0.87, 0.87} 
\definecolor{cold3}{rgb}{0.54, 0.17, 0.89} 

\definecolor{green1}{rgb}{0.0, 0.5, 0.0} 
\definecolor{green2}{rgb}{0.55, 0.71, 0.0} 
\definecolor{green3}{rgb}{0.0, 0.26, 0.15} 

\title{Reduced Order Modeling using Shallow ReLU Networks with Grassmann Layers}

\author[1]{Kayla Bollinger}

\author[1]{Hayden Schaeffer}
\affil[1]{ Department of Mathematical Sciences\\
  Carnegie Mellon University\\
  Pittsburgh, PA, USA  }

\begin{document}
\date{\ }
\maketitle

\begin{abstract}
This paper presents a nonlinear model reduction method for systems of equations using a structured neural network. The neural network takes the form of a ``three-layer''  network with the first layer constrained to lie on the Grassmann manifold and the first activation function set to identity, while the remaining network is a standard two-layer ReLU neural network. The Grassmann layer determines the reduced basis for the input space, while the remaining layers approximate the nonlinear input-output system. 
The training alternates between learning the reduced basis and the nonlinear approximation, and is shown to be more effective than fixing the reduced basis and training the network only. An additional benefit of this approach is, for data that lie on low-dimensional subspaces, that the number of parameters in the network does not need to be large. We show that our method can be applied to scientific problems in the data-scarce regime, which is typically not well-suited for neural network approximations.  Examples include reduced order modeling for nonlinear dynamical systems and several aerospace engineering problems.
\end{abstract}

\section{Introduction} 
\label{sec:intro}

 Deep neural networks (DNN) are a popular approximation technique for problems in image processing, computer vision, natural language processing, and other data-based applications. Their popularity and success is due to their high accuracy, in particular, when trained with a sufficiently large training set. It has been shown that the expressive power of a neural network is a function of the number of trainable parameters, or equivalently for DNN, the number of layers, and thus one can obtain a given level of accuracy by using a large enough set of parameters \cite{Cybenko1989ApproximationBS,YAROTSKY2017103}. However, this can come with potential issues, especially when applied to problems in scientific computing and high-consequence decision making. Specifically, using a large set of trainable parameters often comes at the cost of longer training times, unnecessary model complexity, and more expensive evaluations.  The increase of complexity and evaluation cost can make the network impractical for applications that require repeated queries, like uncertainty analysis and optimization. In addition, expressiveness itself is not the only requirement for applicability of a model, one must also consider stability, robustness, and interpretability which all depend on the structure and size of the network. 
 
 The goal of this work is to construct a reduced order model (ROM) for approximating high-dimensional functions by incorporating a model reduction layer within a shallow neural network. The hope is to maintain the expressiveness of the neural network, but lower the overall complexity through the introduction of these new layers. Since the cost and complexity often increases dramatically with the dimension of the input space, the so called \textit{curse of dimensionality}, reducing the effective dimension of the initial layer should lead to large gains. In addition, this creates a blend between interpretability (through the model reduction layers) and expressiveness (through the neural network), which can be beneficial for many applications.

ROMs are used to approximate high-dimensional complex systems by simpler, and often interpretable, low-dimensional functions which capture the overall dominant behavior of the high-dimensional system.  Projection-based model reduction techniques often construct a low-dimensional approximation directly on the data. One popular approach is the proper orthogonal decomposition (POD), which learns a low-dimensional subspace using the dominant principal components of the data-matrix \cite{berkooz1993proper,holmes2012turbulence}. Then, the original system of equations are transformed to the reduced-space by projecting onto the low-dimensional subspace. Some other approaches for approximating a reduced basis include: global sensitivity analysis \cite{saltelli2008global}, sliced inverse regression \cite{li1991sliced}, and a compressive sensing based approach from \cite{fornasier2012learning}.  
In many cases, it is advantageous to learn both the reduced basis and the governing equations in the reduced domain. For dynamical systems, the dynamic mode decomposition (DMD) extracts a reduced order model by projecting the data onto a low-dimensional subspace and learning a linear evolution equation on the reduced basis \cite{rowley2009spectral, schmid2010dynamic}. The operator inference technique \cite{peherstorfer2016data} can be used to learn a polynomial governing system on the reduced basis, which can better capture the nonlinear behavior on the reduced space, in particular, when the original high-dimensional function is also a polynomial.

For certain problems in scientific computing, both point queries and gradient information can be made available.  Gradient-based model reduction algorithms often use the dominate eigenspace of the second moment matrix as a way of finding the reduced basis. This approach is called the active subspace method \cite{russi2010uncertainty, Constantine_2014,constantine2015active} and has been applied to the identification and analysis of structures in hypersonic flow \cite{constantine2015exploiting}, transonic flow \cite{lukaczyk2014active}, hydrological models \cite{jefferson2015active,jefferson2017exploring}, and ion batteries \cite{constantine2017time}. Active subspaces can also be used to improve the cost of solving Bayesian inverse problems, see for example \cite{cui2014likelihood, constantine2016accelerating}. If the input space has ambient dimension $m$, then under mild assumptions on the function, the error associated with using the $k$-dimensional active subspace (for $k\leq m$) is given by the tail sum of the eigenvalues from $k+1$ to $m$. Thus, when the eigenvalues of the second moment matrix decay rapidly, the active subspace method will be accurate. These methods depend on estimating the spectra accurately using gradient measurements of the high-dimensional function. To avoid the need for obtaining many accurate measurements, multifidelity methods \cite{lam2020multifidelity} can be used to reduce the overall cost by utilizing both high-fidelity and low-fidelity gradient evaluations. 

There are several challenges related to reduced order modeling.  Consider the scalar case: $f:\R^m \to \R$, where $m\gg1$ and where we have constructed the low-dimensional subspace associated with the matrix $U \in \R^{m\times k}$ and projected the function onto the reduced basis, i.e. $F:\R^k \to \R$ with $F(y) = f(Uy)$, where $y\in \R^k$. Even though the input to $F$ is $k$-dimensional, if we are required to evaluate $f$ directly, then we may not have reduced the overall cost. Instead, one can build a surrogate function $g:\R^k \to \R$ so that $g(U^Tx)$ is an approximation to $f(x)$ over a set of given samples. In this way, an evaluation of $g$ has complexity depending on $k$ and not the ambient dimension $m$  \cite{rowley2009spectral, schmid2010dynamic,peherstorfer2016data,Constantine17,Hokanson_2018}.  In this work we will focus on the data-scarce setting, where the number of samples of the data may be smaller than required to get an accurate approximation to the second moment matrix or needed for an accurate approximation in the ambient dimension. This leads to several potential sources of errors throughout the modeling process. First, inaccuracies in the second moment matrix can be viewed as ``noise'' to the active subspace approximation and thus can lead to an inaccurate approximation of the dominate eigenspace.  Secondly, there are errors coming from the dataset having low-fidelity and limited samples. 
Lastly, in this setting, over-parameterized functions could lead to overfitting and inconsistent results.

\subsection{Contribution of this work} 
\label{sec:contribution}

We propose a joint optimization method for simultaneously learning the reduced basis and the surrogate model from point queries. Since the low-data setting can lead to inaccuracies in the eigenspace of the second moment matrix, rather than using the active subspace method directly, we will use it as an initial guess within our method. To approximate the nonlinear system on the reduced basis, we build a surrogate function parameterized by a two-layer fully connected ReLU network. The two-layer network with the subspace reduction layer can be thought of as a ``three-layer'' network with the first layer constrained to be a low-dimensional projection layer and the first activation function to be identity. Although the data-scarce regime is typically not well-suited for neural networks, since we are jointly optimizing the low-dimensional projection within the network structure, the number of free parameters is far fewer than a full network in the ambient dimension. Thus, this approach potentially avoids some of the generalizability issues associated with over-parameterization on small datasets, at least for the applications in this work. We show empirically that this approach is robust to the sample-size and can be applied to various aerospace engineering problems.

\section{Problem Statement} 
\label{sec:ps}

Given a function $f:\Omega\subseteq\R^m \to \R^n$, where $m\gg1$, the goal is to construct a surrogate model of the form: 
$$f(x)\approx g(U^Tx),$$
where $U\in\R^{m\times k}$ is a matrix that maps the input space to a $k$-dimensional subspace $k<m$ and $g:\R^k\to\R^n$ is the approximation of the function with respect to $k$-dimensional inputs. Therefore, the problem is to approximate $f\approx g \circ U^T$ by learning both $U$ and $g$ from a set of $M$-samples $\left\{ x_\ell \right\}_{\ell=1}^M$ (with respect to a sampling measure $\rho$) in order to find a model that depends on fewer inputs than the ambient dimension.

 For our approach, we represent the lower dimensional function $g$ by a shallow neural network and we construct $U$ by a gradient-based dimensional reduction method (inspired by the active subspace technique). In particular, we write $g=g_\theta$ where $\theta\in \mathbb{R}^{d}$ is the set of trainable parameters that define the shallow neural network and we constrain $U$ to be on the Grassmann manifold $Gr(k,m)$. To learn the parameters $\theta$ and $U$, we solve the following minimization problem:
\begin{align}
\min_{\theta\in\R^{d},\  U\in Gr(k,m)} \ \frac{1}{M} \sum_{\ell=1}^M \norm{f(x_\ell) - g_\theta(U^Tx_\ell)}_2^2 + \lambda \| \theta \|_2^2
\label{eqn:min}
\end{align}
where the first term is the empirical risk and the second term is a regularizer on the parameters of the neural network, with $\lambda>0$. To solve Equation~\eqref{eqn:min}, we use an alternating minimization strategy between minimizing the fit of the network for a fixed matrix $U$ and minimizing over $U$ with fixed parameters $\theta$.  The algorithm is detailed in Section \ref{sec:method}. In Sections \ref{sec:AS}-\ref{sec:approximation}, we describe the construction of $g$ and $U$, and recall some of the theory related to the active subspace methods.

\subsection{Active Subspaces} 
\label{sec:AS}

We first recall some of the theory of active subspaces (see \cite{Constantine_2014}), its application to vector-valued functions, and its connection to our problem.

Consider a domain $\Omega\subseteq\R^m$ (centered at the origin) equipped with a probability density $\rho$, that is
$$
\rho(x) > 0 \text{ for } x\in\Omega,\quad \rho(x) = 0 \text{ for } x\notin\Omega.
$$
Let $f$ be continuously differentiable, square integrable with respect to $\rho$, and let the pairwise products of its partial derivatives also be integrable with respect to $\rho$. Consider the symmetric and positive semidefinite matrix $C\in\R^{m\times m}$ defined by
$$
C := \E\left[ Df^T Df \right]
$$
$$
\text{where} \quad 
(Df(x))_{i,j} = \frac{\partial f_i}{\partial x_j}(x) 
\quad \text{and} \quad 
\left( Df^T(x)Df(x) \right)_{i,j} = \sum_{k=1}^n \frac{\partial f_k}{\partial x_i}(x) \frac{\partial f_k}{\partial x_j}(x)
.
$$
Its eigenvalue decomposition can be written as $C := W \Lambda W^T$
with $\Lambda = \mathrm{diag}(\lambda_1,\dots,\lambda_m)$, where $\lambda_1 \geq \lambda_2 \geq \cdots \geq \lambda_m \geq 0$. Define the block-wise structure as follows: $W_1 \in\R^{m\times k}$, $W_2\in\R^{m\times m-k}$, $\Lambda_1\in\R^{k\times k}$, and $\Lambda_2\in\R^{m-k\times m-k}$ and thus:
\[
W = 
\begin{bmatrix}
	W_1 & W_2
\end{bmatrix}
\quad,\quad
\Lambda = 
\begin{bmatrix}
	\Lambda_1 & \\
	& \Lambda_2
\end{bmatrix}.
\]
As in \cite{Constantine_2014}, we define the rotated coordinates $y\in\R^k$ and $z\in\R^{m-k}$ by
$y = W_1^T x$ and $z = W_2^T x$. We next prove an extension of Lemma 2.2 from \cite{Constantine_2014} related to the rotated coordinates of vector-valued functions.

\begin{lemma}
\label{lem:gradeig}
 The sum of the mean-squared gradients of each $f_i$ with respect to the rotated coordinates $y$ and $z$ satisfy:
	\[
	\sum_{i=1}^n\Expect{\grad_y(f_i)^T\grad_y(f_i)}= \lambda_1 + \cdots + \lambda_k
	\]
	\[
	\sum_{i=1}^n\Expect{\grad_z(f_i)^T\grad_z(f_i)}= \lambda_{k+1} + \cdots + \lambda_m
	\]
	where $\lambda_i$ for $i\in[m]$ are the ordered eigenvalues of  $C = \E\left[ Df^T Df \right]$.
\end{lemma}
The proof of Lemma \ref{lem:gradeig} is in Appendix \ref{app:gradeig}. Since $\lambda_j$ are non-negative and ordered, Lemma \ref{lem:gradeig} provides a comparison between the amount that $f$ varies on each of the subspaces associated to $W_\ell$ for $\ell=1,2$. In particular, if $\lambda_{k+1}$ is relatively small, then $f$ depends mainly on the subspace associated with $W_1$, i.e. $\{y: y = W_1^T x, x\in\Omega\}\subseteq \R^k$. If $\lambda_{k+1} =0$, then $f$ is invariant to the subspace $W_2$. Note that we have assumed that each component of $f=[f_1, \cdots, f_n]^T$ depends jointly on the same lower dimensional subspace. This means that obtaining a reduced order model for the function $f$ can be done simultaneously in all components. If this is not the case, then the model reduction must be applied to each component $f_i$ for $i\in [n]$ separately, i.e. a sequence of parallel optimization problems for $f_i:\Omega\subseteq\R^m \to \R$. The results in this section still hold in that setting, since they degenerate to the one-dimensional case discussed in \cite{Constantine_2014}.
  
For a fixed $U$, the minimizer of the risk:
\begin{align}
\min_{g=[g_1,\cdots, g_n]^T} \ \sum_{i=1}^n\ \int_\Omega (f_i(x) - g_i(U^Tx))^2 \, \rho(x) dx 
\label{eq:risk}
\end{align}
 is given by the conditional expectation: 
\begin{align}
g_i(W_1^T x) = g_i(y) := \Expect{f_i(W_1 y + W_2 z)|y} = \int f_i(W_1 y + W_2 z) \rho(z|y) \, dz
\label{eq:gdef}
\end{align}
where $\rho(z|y)$ is the conditional probability density, see \cite{williams_1991}.
As we see in the following theorem, we have that indeed $f$ is close to  $g \circ W_1^T$ whenever $\lambda_{k+1}$ is small.

\begin{theorem}
\label{thm:condexp}
Assume a given probability density $\rho$ is such that the (probabilistic) Poincar\'{e} inequality with respect to $\rho(z|y)$ holds, that is, assuming that the gradient of $\phi:\R^m\rightarrow \R$ is square integrable, then:
$$
\Expect{\left( \phi - \Expect{\phi|y} \right)^2 \Big| y} \leq c \, \Expect{\norm{\grad_z \phi}^2_2 \Big| y}
$$
for some constant $c>0$ depending on the domain $\Omega$ and $\rho$.
Then the mean-squared error of $g\circ W_1^T$ satisfies
$$
\Expect{\norm{ f - g\circ W_1^T }_2^2} \leq c \, (\lambda_{k+1} + \cdots \lambda_m),
$$
where $\lambda_i$ for $i\in[m]$ are the ordered eigenvalues of  $C = \E\left[ Df^T Df \right]$.
\end{theorem}
\noindent The proof of Theorem \ref{thm:condexp} is in Appendix \ref{app:condexp}. This is a vector-valued version of Theorem 3.1 from \cite{Constantine_2014}. Note that if the eigenvalues decay sufficiently rapidly, then the risk will be small. 

Since we are given a set of $M$-samples $\left\{ x_\ell \right\}_{\ell=1}^M$, we obtain an approximation to $C$ by the Monte Carlo estimate $\tilde{C} \in \R^{m\times m}$ defined by: 
\[
C \approx \tilde{C} := \frac{1}{M} \sum_{\ell=1}^M D{f}(x_\ell)^T D{f}(x_\ell)
,\quad \text{i.e.} \quad 
\tilde{C}_{i,j} = \frac{1}{M} \sum_{\ell=1}^M \left( \sum_{k=1}^n \frac{\partial f_k}{\partial x_i}(x_\ell) \frac{\partial f_k}{\partial x_j}(x_\ell) \right)
\]
where $x_\ell\in\Omega$ are $M$ independent and identically distributed (i.i.d.) samples with respect to the sampling measure $\rho$. To approximate the matrix $\tilde{W}_1$, we apply the eigenvalue decomposition to $\tilde{C}$, i.e. $\tilde{C} = \tilde{W}\Lambda\tilde{W}^T$, where $\tilde{W}_1 \in\R^{m\times k}$, $\tilde{W}_2\in\R^{m\times m-k}$, are the block-matrices as defined earlier in this section. To compute $\tilde{W}$, we use the SVD approach. Let $A \in \R^{m \times nM}$ be defined by: 
\begin{align}
A:= \frac{1}{\sqrt{M}}\left[ D{f}(x_1)^T \cdots D{f}(x_M)^T\right]
.
\label{eqn:A}
\end{align}
whose SVD is given by $A = U'\Sigma' V'^T$. Since $AA^T = \tilde{C}$, we then have that $U' = \tilde{W}$ and thus  $U=\tilde{W}_1$.

\subsection{Network Approximation and Alternating Approach} 
\label{sec:approximation}

We propose the following minimization problem for obtaining a reduced order model $g(U^Tx)$ given a set of $M$-samples $\left\{ x_\ell \right\}_{\ell=1}^M$:
\begin{align}
\min_{\theta\in\R^{d},\  U\in Gr(k,m)} \ \frac{1}{M} \sum_{\ell=1}^M \norm{f(x_\ell) - g_\theta(U^T x_\ell)}_2^2 + \lambda \| \theta \|_2^2.
\label{eq:opt}
\end{align}
To solve this minimization problem, we alternate between the following two subproblems: 
\begin{align*}
\min_{\theta\in\R^{d}}\ \frac{1}{M} \sum_{\ell=1}^M \norm{f(x_\ell) - g_\theta(U^T x_\ell)}_2^2 + \lambda \| \theta \|_2^2 \quad \text{(NN Approximation)}
\end{align*}
and 
\begin{align*}
\min_{U\in Gr(k,m)} \ \frac{1}{M} \sum_{\ell=1}^M \norm{f(x_\ell) - g_\theta(U^T x_\ell)}_2^2 \quad \text{(Subspace Approximation).}
\end{align*}

 First, to approximate the nonlinear function over the subspace, we write $g_\theta$ as a shallow neural network with trainable parameters $\theta\in\R^d$. The shallow neural network takes the form of a two-layer ReLU network:
$$
g_\theta(y) = A_2(\text{ReLU}(A_1y+b_1)))+b_2
$$
where the first fully connected linear layer is defined by the matrix $A_1:\R^k\to\R^h$ and bias $b_1\in \R^h$ and the second fully connected linear layer is defined by the matrix $A_2:\R^h\to\R^n$ and bias $b_2\in \R^n$. We use the ReLU activation function: $\left(\text{ReLU}(z)\right)_i = \max(z_i,0)$. The trainable parameter vector $\theta \in \R^d$ is defined as the vector of parameters after concatenating all elements of $A_1$, $A_2$, $b_1$, and $b_2$.  The total number of learnable parameters is $d =h(k + n + 1) + n$.

In Section~\ref{sec:AS}, it was shown that the minimizer over all functions $g$ is obtained by the conditional expectation; however, approximating the conditional expectation (e.g. via the Monte Carlo approximation), would not necessarily be beneficial since the approximation would require evaluations in the ambient dimension $m$. Instead, we want to provide a nonlinear surrogate model in the smaller dimension $k<m$, whose complexity depends on $k$ not $m$. This is one motivation for learning the model through regression.

The other reason to use regression is to deal with the various sources of noise. In the applications discussed in Section \ref{sec:intro}, the noise can arise from multiple sources involving computational errors or inaccuracies. The first is that the underlying function $f$ may not be invariant to the subspace $\{z: z = W_2^T x, x\in\Omega\}\subseteq \R^{m-k}$ (or in a related sense, the decay of the eigenvalues from Theorem \ref{thm:condexp} is slow) and thus the model may incur ``noise'' from the information lost during the subspace reduction. Other sources of noise can come from the training data itself. 
For example, if the data pair $(x,f(x))$ is obtained from a numerical simulation of some complex system with modeling parameter $x$, then $f(x)$ is only an approximation to some underlying function. The numerical accuracy (or inaccuracy in this case) would appear as ``noise'' in the surrogate approximation. Additionally, when using a small sample set, which is the case for some of the experiments in Section \ref{sec:experiments}, the error in approximating the data distribution (for example, the error between $C$ and $\tilde{C}$) can appear as ``noise''. And lastly, if the data is obtained from an experiment or real-world observations, then measurement and/or acquisition noise may be present. This motivates the use of regularized regression, and in particular, the utilization of shallow neural networks to simultaneously fit the data and denoise the system. 

 To obtain $U$, we fix $g_\theta$ and then optimize over the Grassmann manifold $Gr(k,m)$--a compact manifold of all $k$-dimensional linear subspaces in $\R^m$. We use $Gr(k,m)$ since we are only concerned with obtaining a $k$-dimensional subspace, which is represented by the matrix $U$, and not with the representation itself (i.e. the choice of coordinates).  When gradient information is available, we use the procedure described by Equation~\eqref{eqn:A} to obtain an approximation to the active subspace $\tilde{W}_1$ and use it as the initial guess to $U$. When we only have function evaluations (and are unable to obtain the active subspace), then a random initialization scheme is used. An experimental comparison between different initializations is provided in Section \ref{sec:spiral}. In general, approximations that were trained using the active subspace as the initializer perform better (in terms of accuracy/generalizability). The minimization problem in $U$ is nonconvex, and thus a good initializer can be beneficial.

\section{Methodology} 
\label{sec:method}

The optimization of  Equation~\eqref{eq:opt} is done via the alternating minimization approach described in Section~\ref{sec:approximation}. The pseudocode for this algorithm is given in Algorithm \ref{alg:alt}, and is described in detail below. We initialize the trainable parameters of the shallow neural network, i.e. $\theta$,  using the standard Xavier initialization. In particular, each fully connected layer is comprised of weight matrices $A_1\in\R^{h\times k}$, $A_2 \in\R^{n\times h}$ and biases $b_1\in\R^h$, $b_2 \in \R^n$, which are initialized uniformly at random: $(A_1)_{i,j}, (b_1)_i \sim \mathcal{U} \left[-\frac{1}{\sqrt{k}},\frac{1}{\sqrt{k}}\right]$ and  $(A_2)_{i,j}, (b_2)_i \sim \mathcal{U} \left[-\frac{1}{\sqrt{h}},\frac{1}{\sqrt{h}}\right]$.

For the reduced basis subproblem, if Jacobian-information is given, i.e. measurements of $Df(x_\ell)$ over a set of samples, then we initialize $U$ using the active subspace method, that is, $U=W_1$ described in Section \ref{sec:AS}. When the Jacobian is not available,  we use two alternative initialization methods for $U$ (see also \cite{Constantine17}). The first is to initialize $U$ as the identity operator on the first $k$-coordinates of the input space, i.e. 
$$
U = 
\begin{bmatrix}
	I_{k\times k} \\
	0_{m-k\times k}
\end{bmatrix}\
.
$$
The second way is to initialize $U$ at random with the contraint that is remains orthogonal. This is done by creating a random Gaussian matrix of size $m\times k$ whose elements are chosen from the normal distribution with mean zero and standard deviation one, and then setting $U$ to be the orthogonal matrix $Q$ computed via the reduced QR factorization of the random Gaussian matrix.

To optimize Equation~\eqref{eq:opt} we alternate between optimizing over $\theta$, then over $U$. When the active subspace is used as the initial gues, this ordering can lead to dramatic gains in the early training phase. For the NN approximation subproblem, we use the ADAM method \cite{kingma2014adam} with an initial learning rate $\tau$. For the subspace approximation subproblem, we use the Pymanopt package \cite{pymanopt} to solve the Grassmann manifold-constrained least-squares problem using a steepest descent method. We refer to any iteration carried out in these optimization steps as an inner iteration, and we refer to the completion of one-sweep of optimizing over both $\theta$ and $U$ as an outer iteration.

For the NN approximation subproblem (see Section \ref{sec:approximation}), we set a fixed number of $N_\theta$ inner iterations. The number of inner iterations can be set by the users (and thus included as a hyperparameter), or it can be determined by a stopping condition on the loss function for each subproblem. In the experimental results in Section \ref{sec:experiments}, we chose a sufficiently large number of steps, $N_\theta = 5000$, to ensure plateauing of the loss functions for these examples. Some tests showed that early stopping (with a relaxed stopping condition) can reduce the training time while resulting in a similar overall loss (for the full optimization problem). After each outer iteration the learning rate decays, i.e. we reinitialize the hyperparameters in the ADAM method with a reduced learning rate $\tau\leftarrow 0.9\tau$. We set a fixed number of $P$ outer iterations based on empirical tests, in particular,  $P$ should be large enough to allow the loss in Equation~\eqref{eq:opt} to plateau. We define $\theta^{p,n}$ and $U^{p,n}$ to be the computed values of $\theta$ and $U$ after $p$ outer iterations and $n$ (of their respective) inner iterations.  
\begin{algorithm}
\caption{Alternating minimization scheme}
\begin{algorithmic}
\label{alg:alt}
	\STATE Given: $M$ input/output tuples $(x_\ell,f(x_\ell),Df(x_\ell))$
	\STATE Initialize: $\theta\in\R^d$, $U\in\R^{m\times k}$, learning rate = $\tau$, iteration counts $N_\theta$ and $P$
	\WHILE {$p \leq P$}
	\STATE (1) Compute $\theta^{p,N_\theta}$ starting with $\theta^{p,0}$ and applying the ADAM method to:
			\begin{align*}
				\min_{\theta\in\R^{d}} \ \frac{1}{M} \sum_{\ell=1}^M \norm{f(x_\ell) - g_\theta((U^{p,0})^T x_\ell)}_2^2 + \lambda \| \theta \|_2^2
			\end{align*}
		    for $N_\theta$ steps. Then update $\theta^{p+1,0}=\theta^{p,N_\theta}$.
    \STATE (2) Compute $U^{p,N_U}$ as the solution to the Grassmann manifold-constrained least squares problem:
			\begin{align*}
				\min_{U\in Gr(k,m)} \  \frac{1}{M} \sum_{\ell=1}^M \norm{f(x_\ell) - g_{\theta^{p+1,0}}(U^Tx_\ell)}_2^2 .
			\end{align*}
            Then update $U^{p+1,0}=U^{p,N_U}$.
	\STATE (3) Update: $\tau = 0.9 \tau$

\ENDWHILE
\end{algorithmic}
\end{algorithm}

\section{Experimental Results and Applications} 
\label{sec:experiments}

The model contains several hyperparameters, which we list here: the dimension of the reduced basis $k$, the hidden dimension of the shallow network $h$, the regularization weight $\lambda$, and the learning rate  $\tau$. For simplicity, let $X_{Train}$, $X_{Validation}$, and $X$ be the training, validation, and entire available dataset (respectively) and we define their cardinalities by $\abs{X_{Train}}$, $\abs{X_{Validation}}$, and $\abs{X}$.

\subsection{Experiment 1: $U$ initialization comparison.}
\label{sec:spiral}
   We provide a comparison between the various initializers for $U$. In particular, we show that using an approximation to the active subspace, $W_1$, as our initial guess for  $U$ will produce an overall small loss at the end of the training, as compared to the other initializers. We compare the three initializers for $U$ that were discussed in Section \ref{sec:method} on the problem of learning a reduced order model for a nonlinear dynamical system. In particular, define the function $f:\R^3\to\R^3$ by
	$$
	f(x) = 
	\begin{bmatrix}
		x_2^3\\	
		-\left( \frac{x_1 + x_3}{2} \right)^3 - \frac{1}{5}x_2\\
		x_2^3
	\end{bmatrix}
	$$
	and the dynamical system to be the autonomous differential equation $\frac{d}{dt} x(t) = f(x(t))$. This example is related to the sparsity-promoting and data-driven methods for learning unknown governing equations from data, see for example \cite{brunton2016discovering, schaeffer2017learning, schaeffer2017learning, rudy2017data, raissi2018multistep,wu2019numerical, sun2020neupde}. However, in this work, we would like to learn both a reduced basis and a nonlinear governing system approximation, as done in the DMD methodology and in \cite{peherstorfer2016data}.
	
	In this toy example, the system can be reduced to $f(x) = g(U^Tx) = g(y)$ where
	$$
	U = 
	\begin{bmatrix}
		\frac{1}{\sqrt{2}} & 0 \\
		0 & 1 \\
		\frac{1}{\sqrt{2}} & 0
	\end{bmatrix}
	,\quad
	g(y) = 
	\begin{bmatrix}
		y_2^3\\
		-\left( \frac{y_1}{\sqrt{2}} \right)^3 - \frac{1}{5}y_2\\
		y_2^3
	\end{bmatrix}
	$$
	The dataset was constructed using a randomized approach, similar to \cite{schaeffer2018extracting,schaeffer2020extracting}. The dataset is generated by collecting 500 trajectories starting from a uniformly random initial state $x^{0}$ centered at $\tilde{x}^0=[4,3,-2]^T$ with width 2, i.e. $x^{0} \in \mathcal{U}[\tilde{x}^0 - 2, \tilde{x}^0 + 2 ]$. The trajectories were generated using the RK45 method with 202 equally spaced time-stamps over the time interval $[0,5]$. The Jacobian can be extracted directly from $f(x)$. The entire data set $X$  consists of $\abs{X} = 101,000$ samples. For training our model, we use 150 trajectories for our training set $X_{Train}$ and 30 trajectories for our validation set $X_{Validation}$ (i.e. $\abs{X_{Train}} = 30,300$, $\abs{X_{Validation}} = 6,060$). We used a batch size of 16 trajectories in the NN approximation subproblem. The hyperparameters for our model were set to: $k = 2$, $h = 8$, $\lambda = 10^{-7}$, and $\tau = 10^{-3}$.
	
\begin{figure}[t]
        \centering
            \begin{tikzpicture}
                \begin{axis}[
                	xlabel=Epochs, xmin=500, xmax=100000,
                	ylabel=RelError, ymode=log, ymin=0.3, ymax=5,
                	legend style={at={(1.2,1)},anchor=north},
                	title={Initialization of U}]
                \addplot[draw=none, name path=name1, forget plot] table [x=epochs, y=25ptile, col sep=comma] 
                {results/exp1_20/id/val_loss_stats.csv};
                \addplot[color=cold1, very thick] table [x=epochs, y=50ptile, col sep=comma] 
                {results/exp1_20/id/val_loss_stats.csv};
                \addlegendentry{identity};
                \addplot[draw=none, name path=name2, forget plot] table [x=epochs, y=75ptile, col sep=comma] 
                {results/exp1_20/id/val_loss_stats.csv};
                \addplot[color=cold1, forget plot, opacity=0.25] fill between[of=name1 and name2];
                \addplot[draw=none, name path=name1, forget plot] table [x=epochs, y=25ptile, col sep=comma] 
                {results/exp1_20/rn/val_loss_stats.csv};
                \addplot[color=green1, very thick] table [x=epochs, y=50ptile, col sep=comma] 
                {results/exp1_20/rn/val_loss_stats.csv};
                \addlegendentry{random};
                \addplot[draw=none, name path=name2, forget plot] table [x=epochs, y=75ptile, col sep=comma] 
                {results/exp1_20/rn/val_loss_stats.csv};
                \addplot[color=green1, forget plot, opacity=0.25] fill between[of=name1 and name2];
                \addplot[draw=none, name path=name1, forget plot] table [x=epochs, y=25ptile, col sep=comma] 
                {results/exp1_20/acs/val_loss_stats.csv};
                \addplot[color=cold3, very thick] table [x=epochs, y=50ptile, col sep=comma] 
                {results/exp1_20/acs/val_loss_stats.csv};
                \addlegendentry{AS};
                \addplot[draw=none, name path=name2, forget plot] table [x=epochs, y=75ptile, col sep=comma] 
                {results/exp1_20/acs/val_loss_stats.csv};
                \addplot[color=cold3, forget plot, opacity=0.25] fill between[of=name1 and name2];
                \end{axis}
            \end{tikzpicture}
        \caption{Experiment 1: The plot compares the relative error over the training epochs for the three initialization schemes described in Section \ref{sec:method}: identity initialization (in blue), random initialization (in green), and the active subspace initialization (in purple). The median value over 100 trials is displayed as the solid line, and the shaded region encloses the $25^{th}$ to $75^{th}$ percentile. Note that the active subspace initialize reduces the error faster than the other schemes.}
        \label{Fig1}
    \end{figure}
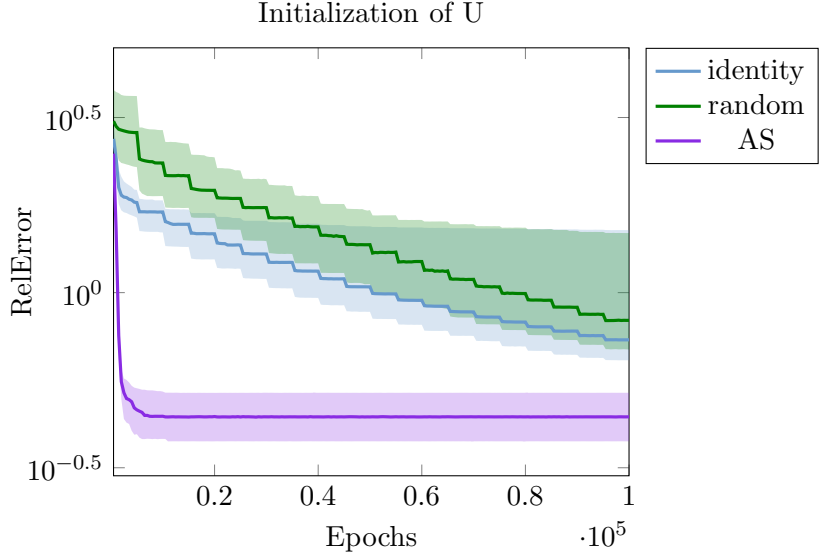

Using 100 randomized trials for training our model, we record our results in Figure~\ref{Fig1}. In Figure~\ref{Fig1}, we plot the epochs (which are the inner iterations for the NN approximation subproblem over the entire training process) versus the relative error, which we define as:
	$$
	\text{RelError}_{p,n}  = \left( \frac{\frac{1}{\abs{X_{Validation}}} \sum_{x\in X_{Validation}} \norm{ f(x)-g_{\theta^{p,n}}((U^{p,0})^T x)}_2^2}{\frac{1}{\abs{X}} \sum_{x\in X} \norm{f(x)}_2^2} \right)^{\frac{1}{2}}
	.
	$$
	The mean-squared errors are normalized separately by the number of points.
Comparing the identity and random initializations, which do not use Jacobian-information, we see that the methods have similar relative errors and will plateau at roughly the same relative error (taking another $10^5$ epochs). The active subspace initialization leads to more dramatic gains in the initial training process, and overall faster convergence. Note that there are still variations on the reduced basis in the training process, i.e. the final subspace may not directly agree with the active subspace that was used to initialize the training. In all cases, the `jumps' in the error occur due to the new estimate of $U^{p,0}$. The jumps in the solid purple curve in Figure~\ref{Fig1} indicate that the alternating minimization leads to a lower relative error than just using the active subspace as the reduced basis. 

\subsection{Experiment 2: Comparing methods using NACA0012.}
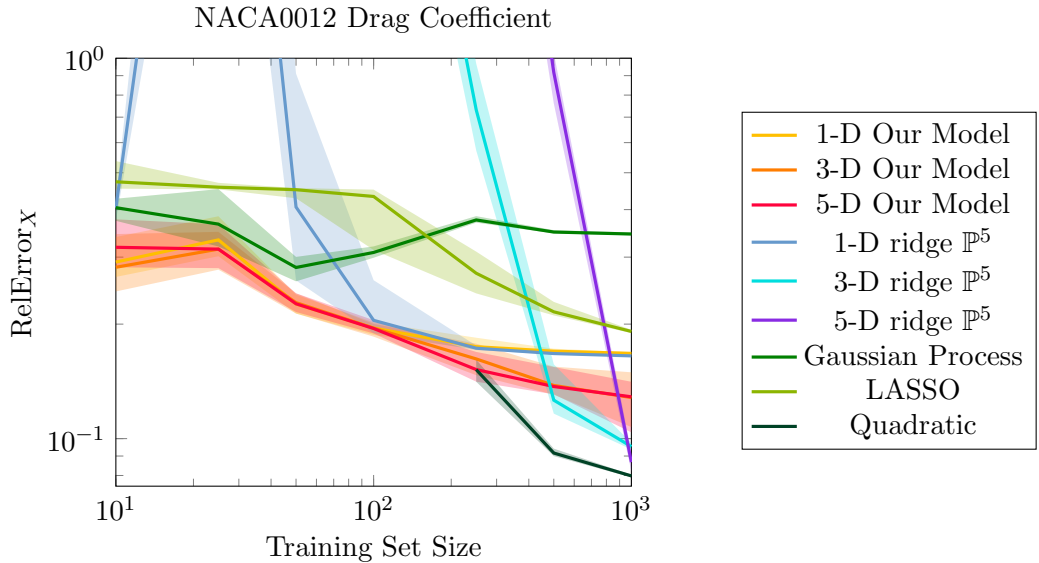
\begin{figure}
        \centering
            \begin{tikzpicture}
                \begin{loglogaxis}[
                	xlabel=Training Set Size, xmin=10, xmax=1000,
                	ylabel=$\text{RelError}_X$, ymin=0.075, ymax=1,
                	legend style={at={(1.5,0.875)},anchor=north},
                	title={NACA0012 Drag Coefficient}]
                
                \addplot[draw=none, name path=name1, forget plot] table [x=num_data, y=25ptile, col sep=comma] {results/exp2/NACAdrag/ourMethod/acsDim_1_scores.csv};
                \addplot[color=warm1, very thick] table [x=num_data, y=50ptile, col sep=comma] {results/exp2/NACAdrag/ourMethod/acsDim_1_scores.csv};
                \addlegendentry{1-D Our Model};
                \addplot[draw=none, name path=name2, forget plot] table [x=num_data, y=75ptile, col sep=comma] {results/exp2/NACAdrag/ourMethod/acsDim_1_scores.csv};
                \addplot[color=warm1, forget plot, opacity=0.25] fill between[of=name1 and name2];
                
                \addplot[draw=none, name path=name1, forget plot] table [x=num_data, y=25ptile, col sep=comma] {results/exp2/NACAdrag/ourMethod/acsDim_3_scores.csv};
                \addplot[color=warm2, very thick] table [x=num_data, y=50ptile, col sep=comma] {results/exp2/NACAdrag/ourMethod/acsDim_3_scores.csv};
                \addlegendentry{3-D Our Model};
                \addplot[draw=none, name path=name2, forget plot] table [x=num_data, y=75ptile, col sep=comma] {results/exp2/NACAdrag/ourMethod/acsDim_3_scores.csv};
                \addplot[color=warm2, forget plot, opacity=0.25] fill between[of=name1 and name2];
                
                \addplot[draw=none, name path=name1, forget plot] table [x=num_data, y=25ptile, col sep=comma] {results/exp2/NACAdrag/ourMethod/acsDim_5_scores.csv};
                \addplot[color=warm3, very thick] table [x=num_data, y=50ptile, col sep=comma] {results/exp2/NACAdrag/ourMethod/acsDim_5_scores.csv};
                \addlegendentry{5-D Our Model};
                \addplot[draw=none, name path=name2, forget plot] table [x=num_data, y=75ptile, col sep=comma] {results/exp2/NACAdrag/ourMethod/acsDim_5_scores.csv};
                \addplot[color=warm3, forget plot, opacity=0.25] fill between[of=name1 and name2];
                
                \addplot[draw=none, name path=name1, forget plot] table [x=samples, y=25ptile, col sep=comma] {results/exp2/NACAdrag/otherMethods/poly5_acs1_drag.csv};
                \addplot[color=cold1, very thick] table [x=samples, y=50ptile, col sep=comma] {results/exp2/NACAdrag/otherMethods/poly5_acs1_drag.csv};
                \addlegendentry{1-D ridge $\mathbb{P}^5$};
                \addplot[draw=none, name path=name2, forget plot] table [x=samples, y=75ptile, col sep=comma] {results/exp2/NACAdrag/otherMethods/poly5_acs1_drag.csv};
                \addplot[color=cold1, forget plot, opacity=0.25] fill between[of=name1 and name2];
                
                \addplot[draw=none, name path=name1, forget plot] table [x=samples, y=25ptile, col sep=comma] {results/exp2/NACAdrag/otherMethods/poly5_acs3_drag.csv};
                \addplot[color=cold2, very thick] table [x=samples, y=50ptile, col sep=comma] {results/exp2/NACAdrag/otherMethods/poly5_acs3_drag.csv};
                \addlegendentry{3-D ridge $\mathbb{P}^5$};
                \addplot[draw=none, name path=name2, forget plot] table [x=samples, y=75ptile, col sep=comma] {results/exp2/NACAdrag/otherMethods/poly5_acs3_drag.csv};
                \addplot[color=cold2, forget plot, opacity=0.25] fill between[of=name1 and name2];
                
                \addplot[draw=none, name path=name1, forget plot] table [x=samples, y=25ptile, col sep=comma] {results/exp2/NACAdrag/otherMethods/poly5_acs5_drag.csv};
                \addplot[color=cold3, very thick] table [x=samples, y=50ptile, col sep=comma] {results/exp2/NACAdrag/otherMethods/poly5_acs5_drag.csv};
                \addlegendentry{5-D ridge $\mathbb{P}^5$};
                \addplot[draw=none, name path=name2, forget plot] table [x=samples, y=75ptile, col sep=comma] {results/exp2/NACAdrag/otherMethods/poly5_acs5_drag.csv};
                \addplot[color=cold3, forget plot, opacity=0.25] fill between[of=name1 and name2];

                \addplot[draw=none, name path=name1, forget plot] table [x=samples, y=25ptile, col sep=comma] {results/exp2/NACAdrag/otherMethods/gaussian_process_drag.csv};
                \addplot[color=green1, very thick] table [x=samples, y=50ptile, col sep=comma] {results/exp2/NACAdrag/otherMethods/gaussian_process_drag.csv};
                \addlegendentry{Gaussian Process};
                \addplot[draw=none, name path=name2, forget plot] table [x=samples, y=75ptile, col sep=comma] {results/exp2/NACAdrag/otherMethods/gaussian_process_drag.csv};
                \addplot[color=green1, forget plot, opacity=0.25] fill between[of=name1 and name2];
                
                \addplot[draw=none, name path=name1, forget plot] table [x=samples, y=25ptile, col sep=comma] {results/exp2/NACAdrag/otherMethods/LASSO_drag.csv};
                \addplot[color=green2, very thick] table [x=samples, y=50ptile, col sep=comma] {results/exp2/NACAdrag/otherMethods/LASSO_drag.csv};
                \addlegendentry{LASSO};
                \addplot[draw=none, name path=name2, forget plot] table [x=samples, y=75ptile, col sep=comma] {results/exp2/NACAdrag/otherMethods/LASSO_drag.csv};
                \addplot[color=green2, forget plot, opacity=0.25] fill between[of=name1 and name2];
                
                \addplot[draw=none, name path=name1, forget plot] table [x=samples, y=25ptile, col sep=comma] {results/exp2/NACAdrag/otherMethods/quadratic_drag.csv};
                \addplot[color=green3, very thick] table [x=samples, y=50ptile, col sep=comma] {results/exp2/NACAdrag/otherMethods/quadratic_drag.csv};
                \addlegendentry{Quadratic};
                \addplot[draw=none, name path=name2, forget plot] table [x=samples, y=75ptile, col sep=comma] {results/exp2/NACAdrag/otherMethods/quadratic_drag.csv};
                \addplot[color=green3, forget plot, opacity=0.25] fill between[of=name1 and name2];
                
                \end{loglogaxis}
            \end{tikzpicture}
        \caption{Experiment 2: Comparison of various methods for approximating the NACA0012 drag coefficient with respect to 18 shape parameters. Our method is labelled as ``k-D Our Model'' where the reduced basis dimensions are $k=1,3,5$. The ``k-D ridge'' curves correspond to the degree-5 polynomial ridge regression model from
    \cite{Hokanson_2018} using active subspaces of dimensions $k=1,3,5$. We also compare to the Gaussian process regression \cite{gpml,sklearn},  
the LASSO problem with a dictionary of monomials up to degree 3 \cite{sklearn}, and a standard quadratic regression problem.  The median relative error is displayed as the solid curves, and the shaded region encloses the $25^{th}$ to $75^{th}$ percentile (using 100 random trials).}
        \label{fig:exp2}
    \end{figure}

This experiment demonstrates the robustness of our model in the scarce-data setting as compared to other approaches. In this test, we find a reduced basis and surrogate model for the drag coefficient associated with the NACA0012 airfoil with respect to 18-shape parameters. The drag coefficient was computed using Stanford University Unstructured (SU2) computational fluid dynamics code \cite{SU2,Hokanson_2018}. The results and comparisons are plotted in  Figure~\ref{fig:exp2}. We compare our model to the degree-5 polynomial ridge regression model from
    \cite{Hokanson_2018} using subspaces of dimensions one, three, and five. In addition, we compare our approach with a Gaussian process regression \cite{gpml,sklearn},  
the LASSO problem (with cross-validation) with a dictionary of monomials up to degree 3 \cite{sklearn}, and a standard quadratic regression problem. The Gaussian process, LASSO, and quadratic regression do not perform model reduction on the input parameters. For each model, we ran 100 trials using  randomly chosen data points for the training set using the sizes $\{10,25,50,100, 250, 500, 1000\}$ (except for the quadratic model, in which we only use $\{250, 500, 1000\}$ since it becomes ill-conditioned for smaller sets). The quadratic model can be used as a baseline error for this problem. For each trial, we calculate their relative errors over the entire dataset $X$, which is defined by: 
	$$
	\text{RelError}_X =  
	\left(
	\frac{\sum_{x\in X} \norm{f(x)-\tilde{f}(x)}_2^2}{\sum_{x\in X}\norm{ f(x)}_2^2} 
	\right)^{\frac{1}{2}}
	$$
	where $\tilde{f}$ represents the trained model for a given approach. 	For our model, we used $P=10$ outer iterations, a batch size of 16 in the NN approximation subproblem, and the remaining hyperparameters were set to: $h = 8$, $\lambda = 10^{-7}$, and $\tau = 10^{-3}$. 
	
	We see that our model performs better than the other approaches for small training sets.  As we increase the training set size, the reduced basis of dimensions 3 and 5 begin to outperform the 1D model. The LASSO model is also robust, but does not reduce the input dimension since sparsity is imposed in the representation (monomials) themselves and not the input space. Note that our 5D network approximation uses 57 free parameters while the 5D polynomial ridge regression approximations uses 252.

\subsection{Experiment 3: Dependence on Basis Dimension and Hidden Dimension.}
\label{sec:NACAhyp}

The number of free parameters in our model depends on the dimension of the reduced basis, denoted by  $k$, and the hidden dimension of the two-layer neural network, denoted by $h$. In this experiment, we apply our method to the NACA0012 airfoil example and measure the error as a function of various values for both $k$ and $h$. Specifically, we vary $k\in\{1,2,3\}$ and $h\in\{8,64,256\}$ over 100 trials with randomly chosen data points, and compared the results in the first table in Table~\ref{tab:exp2}. All other hyperparameters are fixed: $P=10$ outer iterations were used, the training and validation set size was set to 50, the batch size was 16, $\lambda = 10^{-7}$, and learning rate $\tau = 10^{-3}$. We see that as the subspace dimension increases, the number of hidden layers needed to resolve the function also increases. Outside of the case $(k,h)=(1,256)$, where over-fitting may be occurring, the results seem to improve as $h$ increases.

\subsection{Experiment 4: Comparison with a Bowtie Network.}
\label{sec:bowtie}

	A similar network structure can be achieved by using a bowtie network. We compare our model to a shallow bowtie neural network model defined by $\tilde{f}_{\tilde{\theta}}:\R^m\to\R^n$, where
    $$
    \tilde{f}_{\tilde{\theta}}(x) = A_2(\text{ReLU}(A_1(A_0 x) +b_1)))+b_2.
    $$
   The main difference between the bowtie network and our model is that the Grassmann layer is replaced by a fully connected linear layer with $A_0:\R^m\to\R^k$ and without a bias term. The remaining layers are defined as described in Section \ref{sec:approximation}: the first fully connected linear layer is defined by the matrix $A_1:\R^k\to\R^h$ and bias $b_1\in \R^h$ and the second fully connected linear layer is defined by the matrix $A_2:\R^h\to\R^n$ and bias $b_2\in \R^n$. The total number of trainable parameters, i.e. the size of $\tilde{\theta}\in\ R^{\tilde{d}}$, is $\tilde{d} =km+h(k + 1) + n(h+1)$. Note that this is larger than the number of free parameters in our model, since we have the added constraint. We compare the bowtie model with the same set of dimensional parameters: $k\in\{1,2,3\}$ and $h\in\{8,64,256\}$. The bowtie model was optimized over 50,000 epochs (the same total number of epochs/inner iterations used in the NN approximation subproblem for the experiment in Section \ref{sec:NACAhyp}), with the remaining hyperparameters were fixed: the training and validation set size was 50, the batch size was 16, $\lambda = 10^{-7}$, and $\tau = 10^{-3}$.
We ran 100 trials using randomly chosen data points, and the results are summarized in the second table in Table~\ref{tab:exp2}.

The bowtie network produces about $1.2\times$
larger relative errors than our model. For fixed $k$, we see that in Table~\ref{tab:exp2}, as $h$ increases the bowtie model can produce worse results. This is likely due to the lack of structure being imposed onto the system. The diagonal elements of the first table indicate that our model's accuracy improves as with the hidden dimension and basis dimension, which is not the case for the bowtie model. This has two benefits; the first is of practical importance, namely, that the user does not have to manually optimize the hyperparameters. The second is in terms of interpretability, the subspace approximation is likely preserving the dominate features and thus producing an overall model which better fits the dataset.

\begin{table}[h!]
\centering
\begin{tabular}{crrrlcrrr}
\multicolumn{4}{c}{Our Model}                                                                                                                 &  & \multicolumn{4}{c}{Bowtie Model}                                                                                                              \\
\multicolumn{1}{l}{}              & \multicolumn{1}{c}{\textbf{h=8}} & \multicolumn{1}{c}{\textbf{h=64}} & \multicolumn{1}{c}{\textbf{h=256}} &  & \multicolumn{1}{l}{}              & \multicolumn{1}{c}{\textbf{h=8}} & \multicolumn{1}{c}{\textbf{h=64}} & \multicolumn{1}{c}{\textbf{h=256}} \\ \cline{2-4} \cline{7-9} 
\multicolumn{1}{c|}{\textbf{k=1}} & \multicolumn{1}{r|}{0.411}       & \multicolumn{1}{r|}{0.411}        & \multicolumn{1}{r|}{0.484}         &  & \multicolumn{1}{c|}{\textbf{k=1}} & \multicolumn{1}{r|}{0.469}       & \multicolumn{1}{r|}{0.474}        & \multicolumn{1}{r|}{0.468}         \\ \cline{2-4} \cline{7-9} 
\multicolumn{1}{c|}{\textbf{k=2}} & \multicolumn{1}{r|}{0.408}       & \multicolumn{1}{r|}{0.373}        & \multicolumn{1}{r|}{0.370}         &  & \multicolumn{1}{c|}{\textbf{k=2}} & \multicolumn{1}{r|}{0.497}       & \multicolumn{1}{r|}{0.481}        & \multicolumn{1}{r|}{0.523}         \\ \cline{2-4} \cline{7-9} 
\multicolumn{1}{c|}{\textbf{k=3}} & \multicolumn{1}{r|}{0.403}       & \multicolumn{1}{r|}{0.374}        & \multicolumn{1}{r|}{0.331}         &  & \multicolumn{1}{c|}{\textbf{k=3}} & \multicolumn{1}{r|}{0.486}       & \multicolumn{1}{r|}{0.508}        & \multicolumn{1}{r|}{0.505}         \\ \cline{2-4} \cline{7-9} 
\end{tabular}
\caption{Experiment 3 and 4: This table contains the relative error in the drag coefficient associated with the NACA0012 airfoil with respect to 18-shape parameters, comparing our model using various dimensional parameters and a bowtie network of the same size. Specifically, the relative error being shown is the median of the minimum relative error calculated over 100 trials. The dimension of the reduced basis is denoted by  $k$ and the hidden dimension of the two-layer neural network is denoted by $h$.}
\label{tab:exp2}
\end{table}

\subsection{Experiment 5: Applications to Model Reduction for ONERA and HyshotII.}
\begin{table}[h]
\begin{center}
\begin{tabular}{crrr}
\multicolumn{4}{c}{ONERA-M6}                                                                                                                  \\
\multicolumn{1}{l}{}              & \multicolumn{1}{c}{\textbf{h=8}} & \multicolumn{1}{c}{\textbf{h=64}} & \multicolumn{1}{c}{\textbf{h=256}} \\ \cline{2-4} 
\multicolumn{1}{c|}{\textbf{k=1}} & \multicolumn{1}{r|}{0.115}       & \multicolumn{1}{r|}{0.121}        & \multicolumn{1}{r|}{0.150}         \\ \cline{2-4} 
\multicolumn{1}{c|}{\textbf{k=2}} & \multicolumn{1}{r|}{0.110}       & \multicolumn{1}{r|}{0.088}        & \multicolumn{1}{r|}{0.096}         \\ \cline{2-4} 
\multicolumn{1}{c|}{\textbf{k=3}} & \multicolumn{1}{r|}{0.113}       & \multicolumn{1}{r|}{0.076}        & \multicolumn{1}{r|}{0.079}         \\ \cline{2-4} 
\end{tabular}
\end{center}
\caption{This table contains the relative error in the drag coefficient associated with the ONERA-M6 wing with respect to 50-shape parameters \cite{ONERA} using our model with hidden dimension $h$ and a reduced basis dimension $k$. Specifically, the relative error shown is the median of the minimum relative error calculated over 100 trials. The dataset sizes are: $\abs{X} = 297$, $\abs{X_{Train}} = 50$, and $\abs{X_{Validation}} = 50$.}
\label{tab:exp5}
\end{table}

In this experiment, we show that our approach can be applied to other datasets. Specifically, we compute a reduced order model for the drag coefficient associated with the ONERA-M6 wing with respect to 50-shape parameters \cite{ONERA}, and for the normalized integral of pressure over the end of the scramjet engine of the HyShot II vehicle with respect to 7-input parameters \cite{hyshot}. For the ONERA-M6 dataset, we have $\abs{X} = 297$, $\abs{X_{Train}} = 50$, and $\abs{X_{Validation}} = 50$. This dataset includes derivative-information, thus we use the active subspace initialization for $U$. As in Experiments 3 and 4 (Sections \ref{sec:NACAhyp} and \ref{sec:bowtie}, respectively), we vary the same set of dimensional parameters: $k\in\{1,2,3\}$ and $h\in\{8,64,256\}$ and measure the relative error. The remaining hyperparameters were set to: $P=10$ outer iterations, a batch size of 16,  a hidden dimension $h=8$, $k=1$, $\lambda = 10^{-7}$, and $\tau = 10^{-3}$. We ran 100 trials using randomly chosen data points, and the results are summarized in Table~\ref{tab:exp5}. We see that for large hidden dimension (i.e. $h=256$) this model begins to overfit, likely due to the size of the training set. However, similar to the results of Experiment 3 (Section \ref{sec:NACAhyp}), the model does improve as the pair $(h,k)$ increases. 

For the HyShotII dataset, we show that a one-dimensional subspace can be used as an accurate approximation for the 7D parameter space as was done in \cite{constantine2015exploiting}. For this dataset, we have $\abs{X} = 52$, $\abs{X_{Train}} = 10$, and $\abs{X_{Validation}} = 10$. This dataset does not include derivative information, so we used the random initialization scheme for $U$. This system has built in ``noise'' due to the possible errors in the numerical solvers used to generate the dataset. Using the hyperparameters: $P=10$ outer iterations, a batch size of 16, a hidden dimension $h=8$, $k=1$, $\lambda = 10^{-7}$, and $\tau = 10^{-3}$, our model produces an approximation with a relative error of $0.211$. Using these hyperparameters, we ran 100 trials using randomly chosen data points. We observed that for the HyshotII test, overfitting was observed when we used $h\geq 64$.  From the various trials, we found that we only need a hidden dimension of $h=8$, since the training set is very small. This also results in a more compact, and possibly more interpretable, approximation.

\section{Conclusion} 
\label{sec:conclusion}

We proposed an approach for constructing reduced order models using shallow neural networks with structured layers. The first fully connected layer is constrained to lie on the Grassmann manifold in order to map the input data onto a low dimensional subspace. Experimental results on the NACA0012 airfoil show that this constrained layer produces more robust results than a fully connected layer with the same dimensions. When gradient-information is available, the method is initialized using the active subspace method, which outperforms the standard initializers. In addition, it was shown that training the reduced basis jointly with the neural network produces smaller error than using a pre-trained, fixed basis. The method is applied to model reduction of nonlinear dynamical systems and aerospace engineering problems. In several examples, even though the number of samples is relatively small, our method is able to produce meaningful and (relatively) accurate surrogate models. Since the initial layer can dramatically decrease the dimension of the input space, this approach is easy to train and has smaller complexity than a neural network in the ambient dimension. This is beneficial in applications where the model must be queried many times. In addition, the mixture of a shallow and low-complexity neural network with the added constrained layers leads to more interpretable models than a standard neural network.

\section*{Acknowledgments}
The authors would like to acknowledge the support of AFOSR, FA9550-17-1-0125 and the support of NSF CAREER grant $\#1752116$.  The authors would like to thank Jeffrey Hokanson for his help.

\bibliographystyle{plain}
\bibliography{bibfile}

\appendix

\section{Proofs}

\subsection{Lemma \ref{lem:gradeig}}
\label{app:gradeig}

\begin{proof}
The proof generalizes the scalar case, see Lemma 2.2 from \cite{Constantine_2014}. For completeness, we prove it here.

For any $i\in[n]$, each component of $f$ can be written as:
$$
f_i (x) = f_i(WW^T x) = f_i(W_1W_1^T x + W_2W_2^T x) = f_i(W_1 y + W_2 z),
$$
and so by the chain rule we have
$$
\grad_y f_i(x) = \grad_y f_i(W_1 y + W_2 z) = W_1^T \grad_x f_i(W_1 y + W_2 z) = W_1^T \grad_x f_i(x).
$$
It follows that $D_y f(x)^T = W_1^T D_xf(x)^T$, and similarly $D_z f(x)^T = W_2^T D_xf(x)^T$. Therefore,
	\begin{align*}
		\sum_{i=1}^n\Expect{\grad_y(f_i)^T\grad_y(f_i)}
		& = \Expect{\trace{D_y (f)^T D_y (f)}}\\
		& = \Expect{\trace{W_1^T D_x(f)^T D_x(f)W_1}}\\
		& = \trace{W_1^T\Expect{D_x(f)^T D_x(f)}W_1}\\
		& = \trace{W_1^TCW_1}\\
		& = \trace{\Lambda_1}\\
		& = \lambda_1 + \cdots + \lambda_k
	\end{align*}
	The same argument can be applied to the $z$ case.
\end{proof}

\subsection{Theorem \ref{thm:condexp}}
\label{app:condexp}

\begin{proof}
    The proof generalizes Theorem 3.1 from \cite{Constantine_2014}, using our result in Lemma \ref{lem:gradeig}. For completeness, we prove it here.
    
    Note that by Equation (\ref{eq:gdef}), we have that the conditional expectation is zero, i.e. $\Expect{f_i - g_i\circ W_1^T | y} = 0$ for each $i\in [n]$. Therefore, the mean-squared error is controlled by:
	\begin{align}
		\Expect{\norm{f - g\circ W_1^T}_2^2} & = \Expect{\Expect{\norm{f - g\circ W_1^T}_2^2|y}} \label{eq:tower1}\\
			& \leq c \, \Expect{\sum_{i=1}^n\Expect{\norm{\grad_z f_i}^2_2 | y}} \label{eq:poinc}\\
			& = c \, \Expect{\sum_{i=1}^n\Expect{{\grad}_z f_i^T \grad_z f_i|y}} \nonumber\\
			& = c \,  \sum_{i=1}^n \Expect{\grad_z f_i^T \grad_z f_i} \label{eq:tower2}\\
			& = c \, (\lambda_{k+1} + \cdots + \lambda_{m}) \label{eq:lem}
	\end{align}
	where (\ref{eq:tower1}) and (\ref{eq:tower2}) are by the tower property of conditional expectation, (\ref{eq:poinc}) is by the Poincar\'{e} inequality, and (\ref{eq:lem}) follows from Lemma \ref{lem:gradeig}.
\end{proof}

\
\end{document}